\pdfoutput=1
\documentclass[11pt]{article}

\usepackage[final]{acl}

\usepackage{times}
\usepackage{latexsym}

\usepackage[T1]{fontenc}
\usepackage[utf8]{inputenc}

\usepackage{microtype}

\usepackage{inconsolata}

\usepackage{graphicx}
\usepackage{amsfonts} 

\usepackage{subcaption} % For subfigures
\usepackage[most]{tcolorbox}
\usepackage{algorithm}
\usepackage{algorithmic}
\usepackage{booktabs}       % professional-quality tables
\usepackage{amsmath}        % equal
\usepackage{tabularx}       % table
\usepackage{multirow}
\usepackage{tikz}
\usepackage{pgfplots}
\usepackage{makecell}
\usepackage{pgfplotstable}
\usepackage{filecontents}
\usepackage{siunitx}
\usepackage{array} 

\usepackage{xcolor}         % colors
\usepackage[dvipsnames]{xcolor}
\definecolor{c1}{HTML}{63b2ee}
\definecolor{c2}{HTML}{F27970}%{c82423}
\definecolor{c3}{HTML}{f8ac8c} %{FFBE7A}
\definecolor{c4}{HTML}{BEB8DC}

\definecolor{c5}{HTML}{ff7500}
\definecolor{c6}{HTML}{1280B0}
\definecolor{c7}{HTML}{17A1DE}
\definecolor{lightblue}{HTML}{53D2FF} % 浅蓝色
\definecolor{lightyellow}{HTML}{F9DE00} % 浅红色

\title{Think Less, Know More: State-Aware Reasoning Compression with Knowledge Guidance for Efficient Reasoning}

\author{
 \textbf{Yi Sui\textsuperscript{1}},
 \textbf{Chaozhuo Li\textsuperscript{2}},
 \textbf{Dawei Song\textsuperscript{1,3} \thanks{Corresponding author.}}
\\
 \textsuperscript{1}Beijing Institute of Technology, China \\
 \textsuperscript{2}Beijing University of Posts and Telecommunications, China \\
 \textsuperscript{3}The Open University, UK
 \\
 \small{
   \textbf{Correspondence:} \href{mailto:email@domain}{dawei.song2010@gmail.com}
 }
}

\begin{document}
\maketitle
\begin{abstract}

% Large Reasoning Models (LRMs) achieve significant performance improvements on complex tasks, benefiting from long Chains of Thought (CoT). However, these models often suffer from the overthinking phenomenon: they persistently follow deep reasoning patterns regardless of task complexity, resulting in excessive reasoning steps and high latency. Existing CoT compression methods generally struggle to balance accuracy and efficiency, and lack fine-grained adaptability to step-level redundancies and reasoning biases. 
% To address these limitations, we propose State-Aware Reasoning Compression with Knowledge Guidance (STACK), a novel framework that dynamically regulates CoT compression at the step level based on reasoning states while integrating a retrieval-augmented strategy. STACK constructs contrastive samples online and adaptively switches between self-prompted and knowledge-guided compression strategies at each step, while an answer-convergence-based early-stopping mechanism suppresses redundant verification. The framework further leverages reward-difference-driven training by combining Proximal Policy Optimization (PPO) and Direct Preference Optimization (DPO), enabling the model to autonomously learn optimal compression strategies. 
% Experiments on three mathematical reasoning benchmarks show that STACK achieves the optimal balance between reasoning length reduction and performance compared to existing methods. reducing the average response length by 59.9\% while improving average accuracy by 4.8.

Large Reasoning Models (LRMs) achieve strong performance on complex tasks by leveraging long Chain-of-Thought (CoT), but often suffer from overthinking, leading to excessive reasoning steps and high inference latency. Existing CoT compression methods struggle to balance accuracy and efficiency, and lack fine-grained, step-level adaptation to redundancy and reasoning bias.
Therefore, we propose State-Aware Reasoning Compression with Knowledge Guidance (STACK), a framework that performs step-wise CoT compression by explicitly modeling stage-specific redundancy sources and integrating with a retrieval-augmented guidance. STACK constructs online long-short contrastive samples and dynamically switches between knowledge-guided compression for uncertain or biased reasoning state and self-prompted compression for overly long but confident state, complemented by an answer-convergence-based early stopping mechanism to suppress redundant verification.
We further propose a reward-difference-driven training strategy by combining Proximal Policy Optimization (PPO) and Direct Preference Optimization (DPO), enabling models to learn state-conditioned compression strategies. Experiments on three mathematical reasoning benchmarks show that STACK achieves a superior accuracy-efficiency balance, reducing average response length by 59.9\% while improving accuracy by 4.8 points over existing methods.

\end{abstract}

\section{Introduction}
Driven by reinforcement learning (RL) with verifiable rewards, Large Reasoning Models (LRMs)~\cite{jaech2024openai, guo2025deepseek} have developed long Chain-of-Thought (CoT) reasoning patterns~\cite{wei2022chain, chen2025towards}. By leveraging deep logical deductions, extensive exploration, and mechanisms for reflection and self-verification, this slow-thinking paradigm substantially enhances reasoning depth and flexibility~\cite{yeo2025demystifying}. It induces a qualitative shift from shallow pattern matching toward structured and verifiable reasoning in challenging tasks such as mathematical proofs, complex programming and logical analysis. However, long CoT reasoning also introduces an overthinking phenomenon: regardless of task complexity, the model consistently follows a deep reasoning pattern, leading to extended reasoning steps that substantially increase response latency and computational cost~\cite{chiang2024over, sui2025stop}.

\begin{figure}
    \centering
    \includegraphics[width=1\columnwidth]{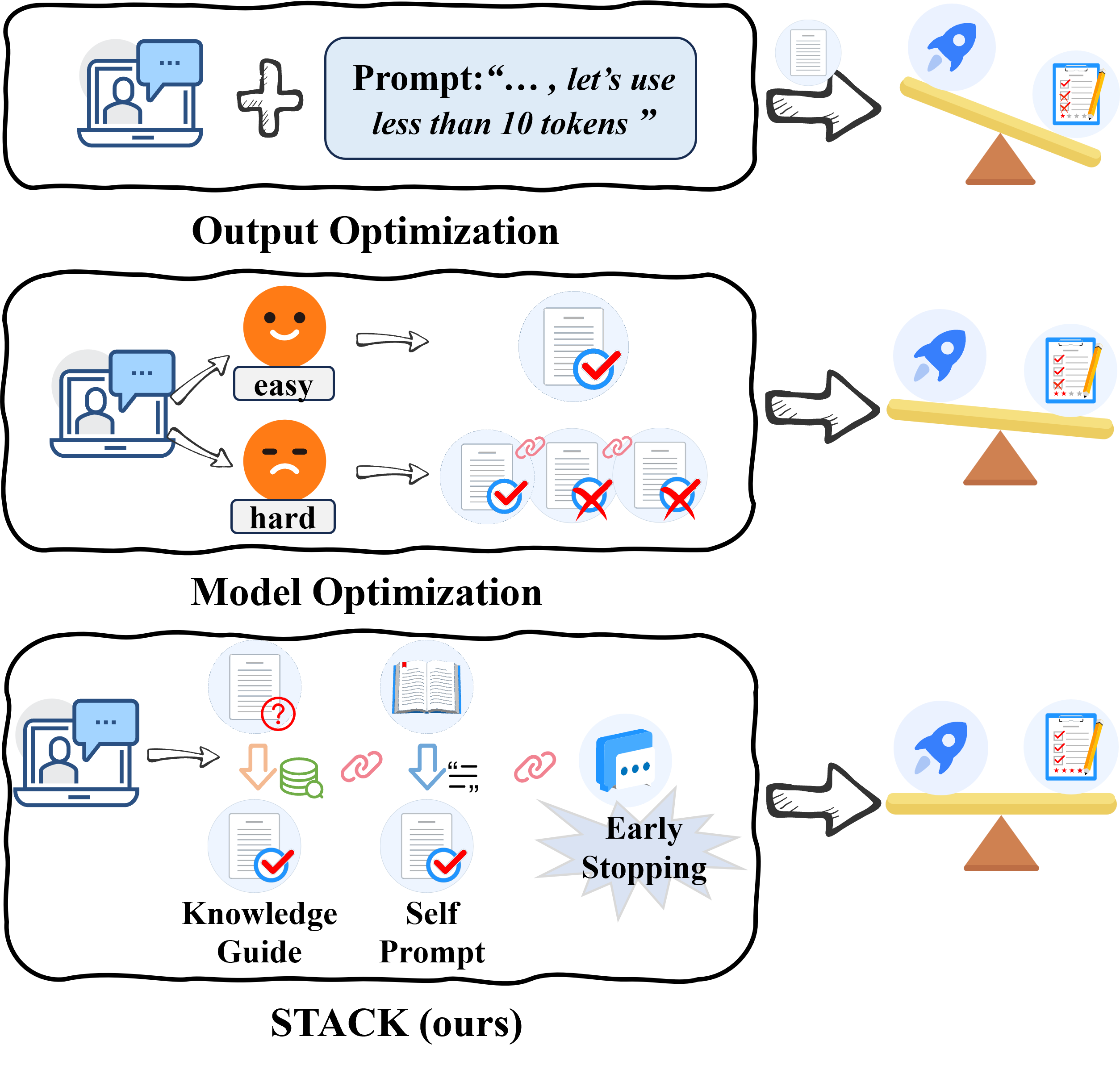}
    \vspace{-7mm}
    \caption{Comparison of CoT Compression Methods.}
    \label{fig:compare}
    \vspace{-18pt}
\end{figure}

To address this issue, CoT compression techniques~\cite{liu2025efficient, qu2025survey} aim to improve reasoning efficiency while preserving performance, primarily through output or model optimization, as shown in Figure~\ref{fig:compare}. \textit{Output optimization} typically relies on prompt design to encourage concise reasoning~\cite{xu2025chain, han2025token}. However, it may compromise reasoning completeness and coherence, and is inherently constrained by performance ceilings. 
In contrast, \textit{model optimization} promotes shorter reasoning by introducing reward functions or penalty terms within RL, to train the models to adaptively decide when to engage explicit reasoning~\cite{zhang2025adaptthink, hou2504thinkprune}. This type of methods are effective in reducing reasoning length at the cost of reasoning accuracy. However, in long CoT, the redundancy arises from heterogeneous sources, leading to varied \textit{reasoning states} at different stages of reasoning, such as \textit{repetitive elaboration under high confidence} and \textit{excessive verification under uncertainty}. Although existing model optimization methods adopt online learning strategies, they primarily adjust compression intensity according to task-level complexity~\cite{yan2025long, fang2025thinkless}. 
The absence of a finer-grained, step-wise modeling of redundancy sources prevents the existing methods from dynamically selecting the appropriate compression strategies required for different reasoning states along the reasoning chain.

% Moreover, recent studies demonstrate that long CoT can lead to error propagation and the risk of hallucination~\cite{lu2025auditing}: early-stage local biases can be continuously amplified through reflection and self-persuasion, ultimately degrading the reliability and accuracy of the reasoning process. 
Moreover, recent studies show that long CoT may aggravate hallucinations~\cite{lu2025auditing}: early-stage local biases are propagated and amplified through reflection and self-persuasion, leading to error accumulation that undermines reasoning reliability and accuracy.
Retrieval-augmented generation (RAG) is a widely used approach to tackle the problem. However, given the extended reasoning chains of LRMs, the external knowledge introduced as contextual input tends to be gradually dominated by internal inference as reasoning unfolds.
This observation motivates a step-wise integration of COT compression and knowledge guidance, enabling timely correction of reasoning deviations while suppressing redundancy.
% This finding highlights the necessity of incorporating external knowledge as an explicit corrective signal to counteract bias accumulation during long-chain reasoning.

% Furthermore, offline model optimization strategies predominantly rely on manually annotated data or concise answers provided by teacher models~\cite{munkhbat2025self, qiao2025concise}, causing models to merely imitate fixed compression patterns. This reliance constrains their ability to autonomously explore input-conditioned optimal compression strategies and prevents real-time intervention during the compression process. 

% To tackle the aforementioned challenges, we propose State-Aware Reasoning Compression with Knowledge Guidance (STACK), a unified framework that dynamically regulates COT compression based on step-level reasoning states, while explicitly integrating retrieval-augmented mechanisms into the compression process. During sampling, STACK constructs contrastive long–short CoT samples online. Guided by step-level reasoning states as decision signals, the framework adaptively switches between self-prompted compression and knowledge-guided compression,
To address these challenges, we propose State-Aware Reasoning Compression with Knowledge Guidance (STACK), a unified framework that dynamically regulates step-level CoT compression while integrating with a retrieval-augmented guidance. During sampling, STACK constructs online contrastive long–short CoT samples and adaptively switches between different compression strategies for different reasoning states: specifically, a knowledge-guided compression for uncertain or biased reasoning state and a self-prompted compression for overly long but confident state, to address stage-specific sources of redundancy. This would
enable the model to learn when to incorporate external knowledge for correction and when to condense redundant reasoning. In addition, we introduce an early stopping mechanism based on the convergence of answer distribution to identify and suppress redundant verification steps, further improving CoT conciseness.

At the training stage, STACK combines the reward modeling capability of Proximal Policy Optimization (PPO)~\cite{schulman2017proximal} with the lightweight contrastive learning paradigm of Direct Preference Optimization (DPO)~\cite{rafailov2023direct}. By driving policy updates through reward differences, the framework enhances the model's ability to autonomously learn state-conditioned optimal compression strategies, while leveraging high-quality external evidence to guide inference under the uncertain states, thereby effectively mitigating reasoning bias and error accumulation.
% Experiments on three mathematical reasoning benchmarks demonstrate that STACK consistently improves both reasoning efficiency and accuracy, achieving a superior trade-off between efficiency and reliability compared to existing CoT compression methods. 

Our work claims the following key innovations:
\begin{enumerate}
  \vspace{-3mm}
  \item We propose STACK, a state-aware framework that dynamically switches targeted compression strategies to address step-level sources of redundancy.
  \vspace{-3mm}
  \item We introduce a knowledge-guided contrastive decoding mechanism with an answer-distribution-based early stopping mechanism, enabling controllable suppression of redundant reasoning while correcting uncertain or biased paths.
  \vspace{-3mm}
  \item Extensive experiments on three mathematical reasoning benchmarks demonstrate that STACK consistently outperforms existing CoT compression methods, achieving a superior balance between reasoning accuracy and efficiency across different model scales.
\end{enumerate}
\vspace{-3mm}

\begin{figure*}
    \centering
    \includegraphics[width=1\textwidth]{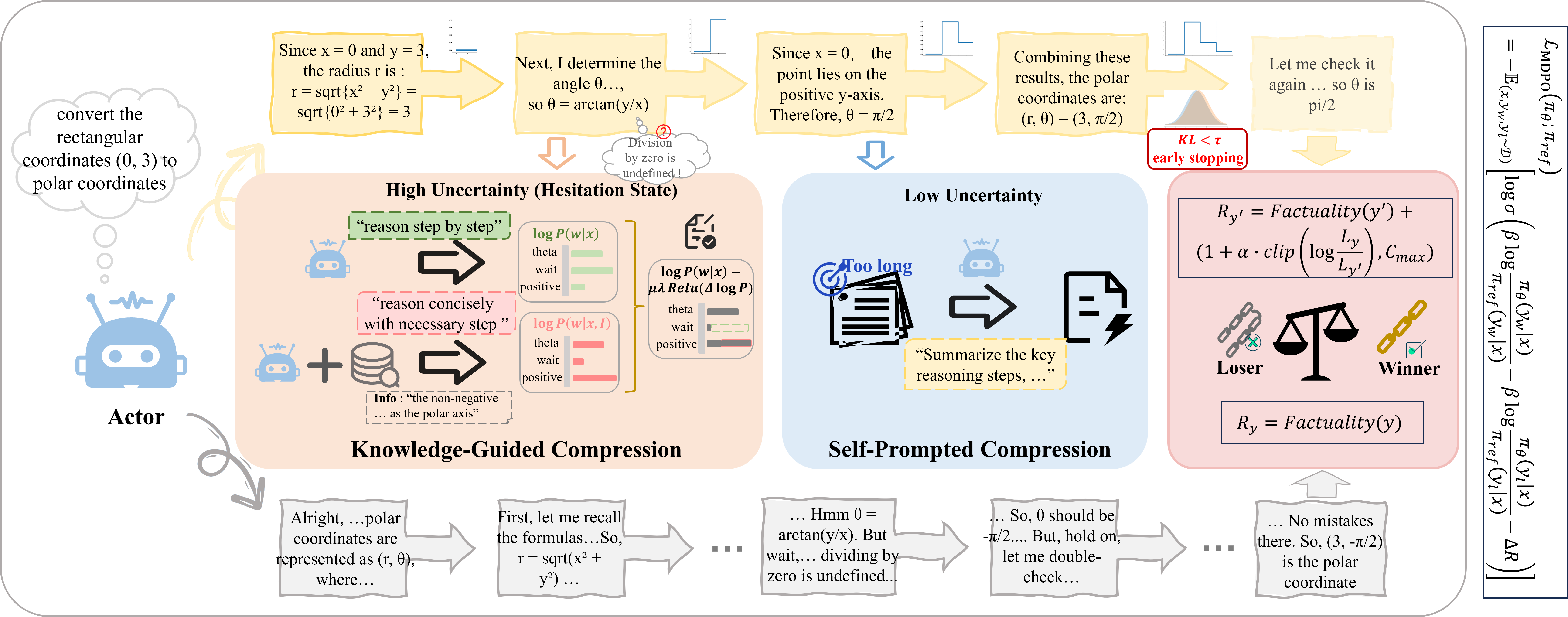}
    \caption{Framework of State-Aware Reasoning Compression with Knowledge Guidance.}
    \label{fig:framework}
    \vspace{-8pt}
\end{figure*}

\section{Related work}

\textbf{Chain-of-Thought Compression.} Existing CoT compression strategies to mitigate overthinking primarily fall into three classes: prompt engineering~\cite{xu2025chain, yan2025inftythink}, which constrains reasoning length or format via explicit prompts; reasoning output optimization~\cite{shen2025codi, cheng2024compressed}, which compresses reasoning steps into a latent space or adjusts strategies during inference; and model optimization~\cite{shen2025dast, luo2025o1}, which enhances reasoning efficiency via online RL or offline fine-tuning. While online RL encourages concise reasoning through reward design or penalty item~\cite{team2025kimi, zhang2025continue}, and offline fine-tuning uses variable-length CoT data~\cite{munkhbat2025self}. However, these methods overlook step-level redundancy causes, and frequently compromise accuracy by ignoring the influence of the model's intrinsic biases. Our work addresses this by dynamically switching compression strategies at the step level and introducing external knowledge to correct reasoning biases.

\textbf{Model Hallucination and Contrastive Decoding.} Hallucination remains a key challenge for LRMs~\cite{huang2024survey, tonmoy2024comprehensive}. Retrieval-Augmented Generation (RAG), which incorporates real-time external knowledge, has emerged as a mainstream approach to address hallucinations~\cite{gao2023retrieval, fan2024survey}.
To ensure models incorporate external knowledge in long-context inputs, contrastive decoding is proposed to subtly strengthen context guidance in reasoning~\cite{li2023contrastive}.
 For instance, context-aware decoding~\cite{shi2024trusting} suppresses hallucinations by contrasting outputs distributions with and without external context, but assumes perfectly reliable knowledge. Recent work integrates both relevant and irrelevant knowledge to improve robustness in conflict scenarios~\cite{2024Enhancing}. By masking retrieval heads to trigger hallucinations, some approaches dynamically contrast the original and masked model outputs via conditional entropy to mitigate hallucinated responses~\cite{2024DeCoRe}. 
%  Unlike these, our Knowledge-Guided Contrastive Decoding employs an entropy-based trust gate to dynamically balance external evidence with parametric knowledge and penalizes redundancy via output distribution divergence, achieving both knowledge enhancement and concise output.
% In contrast, our Knowledge-Guided Contrastive Decoding adaptively integrates external evidence with parametric knowledge and explicitly discourages redundancy, leading to more accurate and concise outputs.
In contrast, our Knowledge-Guided Contrastive Decoding adaptively integrates external evidence with parametric knowledge, with the aim of improving output accuracy while mitigating redundancy.

\section{State-Aware Reasoning Compression with Knowledge Guidance (STACK)}

\subsection{Overview}

% As shown in Figure~\ref{fig:framework}, STACK builds upon the DPO training paradigm to enable adaptive reasoning-chain compression through online long–short contrastive sampling, thereby avoiding the training instability commonly associated with PPO and the prohibitive computational overhead of GRPO.
% For each input, the original complete reasoning chain serves as the long sample, while the short sample is dynamically generated through step-wise segmentation and fine-grained compression.

% To enable step-wise online compression, STACK monitors local information entropy to detect hesitation states, signaling high reasoning uncertainty. Upon detection, STACK switches to a knowledge-guided compression strategy which employs the contrastive decoding to steer the model along evidence-supported reasoning paths. If absent but the response is overly long, self-prompted strategy compresses redundant or repetitive steps. An early stopping mechanism further terminates generation when reasoning stabilizes, preventing redundant verification. 
% Compressed outputs are aggregated into a short chain, forming a contrastive pair with the original long chain.
% Finally, STACK evaluates the quality of both chains using a reward function, incorporating the reward differences into the DPO loss to drive policy updates,  enabling efficient chain-of-thought compression without sacrificing reasoning accuracy. The following sections provide detailed descriptions of each module.

As illustrated in Figure~\ref{fig:framework}, STACK builds upon the DPO paradigm to achieve adaptive chain-of-thought compression via online long–short contrastive sampling, avoiding the training instability of PPO and the high computational cost of GRPO. For each input, the original reasoning chain is treated as the long sample, while the short sample is generated online through step-wise segmentation and fine-grained compression.

STACK enables step-wise compression by monitoring local information entropy to identify hesitation states that indicate high reasoning uncertainty. When such states are detected, a knowledge-guided compression strategy employs contrastive decoding to steer generation toward evidence-supported reasoning. If absent but the response is overly long, a self-prompted compression strategy removes redundant or repetitive steps. An answer-convergence-based early stopping mechanism further suppresses unnecessary verification once reasoning stabilizes.
The resulting compressed reasoning chains are paired with their original counterparts to form contrastive samples. STACK then evaluates both chains with a reward function and integrates the reward difference into the DPO loss, enabling efficient reasoning compression while preserving accuracy. Subsequent sections detail each component.

\subsection{Detection of hesitation state}
Existing research addressing the overthinking issue in LRMs typically relies on fine-tuning with offline-constructed datasets or implements coarse-grained compression based on problem-level complexity~\cite{zhang2025continue}, which limits the flexibility and adaptability of such approaches. In practice, redundant reasoning primarily stems from two sources: first, the presence of substantial expressions that convey little substantive information; second, the model's low confidence in the intermediate reasoning outcome, necessitating iterative verification to confirm correctness~\cite{2025Beyond}. We refer to the latter as the hesitation state.

To enable fine-grained compression, we introduce local entropy during sampling to track the model's reasoning state at each step, allowing dynamic, targeted compression based on the source of redundancy. For the token generated at step $t$ with length $l$, its generation probability distribution is denoted as $P(x)$. The local information entropy is defined as
\vspace{-7pt}
\begin{equation}
H(X_t) = -\frac{1}{l} \sum_{x \in \mathcal{V}} P(x) \log P(x)  \text{,} 
\vspace{-5pt}
\end{equation}
where $\mathcal{V}$ represents the vocabulary space. During reasoning, we further maintain the mean local entropy over a sliding window of size $m$ to model the historical confidence baseline for the current sample. 
% If the local entropy $H(X_t)$ at step $t$ does not exceed the historical mean, the model is considered to be in a relatively high-confidence state, and self-prompted compression is applied to eliminate redundant expressions. 
If $H(X_t)$ significantly exceeds the historical mean, it indicates that the model exhibits abnormal uncertainty at this reasoning step. 
In this case, the model is judged to have entered a hesitation state and the framework activates a knowledge-guided compression strategy, introducing external knowledge to steer the reasoning direction and reduce ineffective repetitive reasoning. More detailed analyses are provided in the appendix~\ref{sec:hesitationAppendix}.

\subsection{Step-level compression strategy}

\textbf{Knowledge-Guided Compression.} 
When the model enters a hesitation state, relying solely on self-guided reasoning tends to amplify local errors or perpetuate incorrect reasoning trajectories, necessitating a knowledge-guided mechanism that simultaneously corrects reasoning direction and suppresses redundancy verification. To this end, we propose Knowledge-Guided Contrastive Decoding (KGCD). KGCD treats retrieved information as a factual anchor and dynamically regulates generation by contrasting the distributional differences between the model's parametric knowledge and external evidence.

To enhance the discriminability of distributional contrasts, KGCD employs a prompt asymmetry strategy at the input stage: the original model $P_{\mathrm{model}}$ maintains its long CoT reasoning behavior, while explicit constraints for concise, evidence-driven reasoning are injected into the system instruction of the retrieval-conditioned model $P_{\mathrm{ctx}}$. This induces a behavioral-level divergence between standard stepwise reasoning and retrieval-guided concise reasoning, forming the basis for fine-grained contrastive decoding in log-probability space. Formally, KGCD modulates the generation log-probability for each token $w$ as follows:
\vspace{-5pt}
\begin{multline}
\log \tilde{P}(w) = \log P_{\mathrm{model}}(w) -  \mu \cdot \lambda_w \cdot \\ \mathrm{ReLU}\Big( \log P_{\mathrm{model}}(w) - \log P_{\mathrm{ctx}}(w) \Big) \text{,}
\vspace{-5pt}
\end{multline}
where $\log P_{\mathrm{model}}(w)$ and $\log P_{\mathrm{ctx}}(w)$ represent the log generation probabilities for the current token from the original model and the knowledge-guided model, respectively.

Due to the prompt asymmetry strategy, when $ \log P_{\mathrm{model}}(w) >> \log P_{\mathrm{ctx}}(w) $, it indicates that the token is overly favored without external evidence support, manifesting as redundant reasoning or unfounded self-expansion. Conversely, when the retrieval context significantly increases the relative probability of certain tokens, it indicates that external knowledge provides a clear direction for reasoning, helping to correct reasoning bias caused by knowledge gaps. Therefore, this probability difference term serves the dual role of redundancy identification and directional guidance. The ReLU term becomes inactive when the knowledge-guided model assigns higher probability to a token, contrastive decoding is realized implicitly at the distribution level by selectively suppressing model-favored but knowledge-inconsistent tokens, thereby reallocating probability mass toward evidence-supported candidates without explicitly boosting their logits.

To dynamically adjust the intervention strength, we design a dynamic trust gate based on the relative entropy reduction: $\lambda_w = \mathrm{clip}\big((H_w^{\mathrm{model}} - H_w^{\mathrm{ctx}}) / (H_w^{\mathrm{model}} + \epsilon), 0, 1\big)$. This gate quantifies the degree to which the retrieval context reduces uncertainty relative to the model itself, thereby assessing the reliability of external knowledge. When the entropy of the retrieval context is significantly lower than the model's entropy, $\lambda_w$ approaches 1, indicating that the external knowledge provides high-confidence guidance, and the system fully activates the compression mechanism. Otherwise, $\lambda_w$ automatically decays to zero to avoid negative intervention from low-quality evidence. Here, $\epsilon=1e-06$ is a smoothing constant to prevent division by zero. Additionally, the hyperparameter $\mu$ serves as a global compression strength coefficient to modulate the suppression magnitude.

Unlike traditional retrieval-augmented generation, which incorporates external knowledge implicitly as contextual prompts, KGCD performs continuous and controllable knowledge injection via distribution-level contrastive decoding. This enables external knowledge to precisely calibrate reasoning direction while effectively suppressing redundant generation at the token level, thereby achieving a superior balance between accuracy and reasoning efficiency.

\textbf{Self-Prompted Compression.} When the model is in a high-confidence reasoning state but the response exceeds the predefined maximum length of 2000 tokens\cite{zhao2025mutis}, STACK activates the self-prompted compression strategy. Under the constraint of  lightweight prompt (e.g., ``Summarize the key reasoning steps, maintaining logical completeness and removing redundant elaboration."), LRMs are encouraged to generate reasoning steps that are both logically complete and concisely expressed based on the current reasoning context. This approach significantly reduces token usage while avoiding information loss caused by coarse-grained global compression.

\subsection{Early stopping mechanism} 
Given that LRMs exhibit introspection and self-verification capabilities, they may continue to generate redundant confirmatory reasoning even after reaching a final answer, resulting in unnecessary computational overhead~\cite{qiao2025concise}. This observation motivates the incorporation of an early stopping mechanism that explicitly detects information saturation during online reasoning, enabling timely termination of redundant steps and facilitating more effective CoT compression as well as more informative training signals.

To this end, we design an early stopping mechanism based on convergence at the answer level. Effective reasoning steps are expected to contribute substantive information toward deriving the final answer. In contrast, when subsequent steps serve only as redundant verification, the model's probability distribution over candidate answers tends to stabilize. Accordingly, we take the reasoning outputs at steps $t-1$ and $t$, denoted as $(a_{t-1}, a_{t})$, as prefix conditions, and compute the conditional probability distributions over the same answer space. The information gain introduced by the additional reasoning step is quantified by the Kullback-Leibler (KL) divergence between these two distributions, formally defined as:
\vspace{-5pt}
\begin{equation}
    \mathrm{IG}_t = \mathrm{KL}(P_t || P_{t-1}) \text{,}
\vspace{-5pt}
\end{equation}
The probability at step $t$ is defined as 
$P_t = P(A \mid a_t, \text{`Therefore the final answer is \string\boxed'})$,
where $P_{t-1}$ follows an analogous definition. When the information gain falls below a predefined threshold $\tau$, it indicates that the current reasoning can no longer significantly alter the model's answer judgment. The reasoning process is therefore regarded as converged, and early stopping is triggered.

\subsection{Reward function}

To jointly optimize reasoning quality and output efficiency during training, we design a reward mechanism that integrates factual accuracy with compression gain. In a complete sampling trajectory, given an input $x$, the intermediate results produced at each reasoning step are compressed and sequentially aggregated to form a shortened CoT $y^\prime$. Together with the original long CoT $y$ generated under the same sampling trajectory, they constitute a paired long–short contrastive sample. 

The reward for the original chain $y$ reflects only its factual accuracy, $\mathcal{R} = \mathrm{Factuality}(y)$, serving as the baseline for subsequent compression effectiveness. The reward for the compressed chain $y^\prime$, however, considers both accuracy and its compression ratio relative to $y$:
\vspace{-5pt}
\begin{multline}
\mathcal{R}_{y^\prime} = \mathrm{Factuality}(y^\prime)\times\\ \left( 1 + \alpha \cdot \min \left( \log\left( \frac{L_y}{L_{y^\prime}} \right), C_{\max} \right) \right)  
\text{,}
\vspace{-5pt}
\end{multline}
where $L_y$ and $L_{y^\prime}$ denote the token lengths of the original and compressed reasoning chains, respectively. The logarithmic term measures compression efficiency, and the clipping function $\min(\cdot, C_{\max})$ constrains the maximum gain ($C_{\max}=1.5$) to prevent obtaining unreasonable rewards through excessive compression. The hyperparameter $\alpha$ balances the accuracy constraint and the compression incentive, ensuring that improvements in efficiency do not compromise reasoning quality.

% This design employs factuality as the dominant signal and incorporates compression benefit as a constrained multiplicative adjustment. It thereby incentivizes more concise reasoning expressions while guaranteeing accuracy, prevents the strategy from degrading to guessing the answer directly, and guides the model to learn efficient yet robust chain-of-thought compression behaviors.
By prioritizing factuality and introducing compression benefit as a constrained multiplicative term, this design promotes concise reasoning without sacrificing accuracy. It avoids degenerate shortcut behaviors and steers the model toward learning efficient and robust chain-of-thought compression.

\begin{table*}[t]
\centering
\small
\setlength{\tabcolsep}{7pt}
\begin{tabular}{c|ccc|ccc|ccc|ccc}
\toprule
\multirow{2}{*}{\textbf{Method}}
& \multicolumn{3}{c|}{\textbf{GSM8K}} 
& \multicolumn{3}{c|}{\textbf{MATH500}} 
& \multicolumn{3}{c|}{\textbf{AIME24}} 
& \multicolumn{2}{c}{\textbf{Average}} \\
\cmidrule(r){2-4} \cmidrule(r){5-7} \cmidrule(r){8-10} \cmidrule(r){11-13}
& \textbf{Acc} & \textbf{Len} & \textbf{TE} 
& \textbf{Acc} & \textbf{Len} & \textbf{TE} 
& \textbf{Acc} & \textbf{Len} & \textbf{TE} 
& $\Delta$\textbf{Acc} & $\Delta$\textbf{Len} & \textbf{Lat (s)} \\
\midrule

\multicolumn{13}{c}{\textit{DeepSeek-R1-Distill-Qwen-1.5B}} \\
% \midrule
Original
& 83.2 & 2247 & 3.70
& 82.4 & 4876 & 1.69
& 29.4 & 16814 & 0.17
& -- & -- & 16.14 \\
Prompt
& 85.3 & 2134 & 4.00
& 83.7 & 4313 & 1.94
& 28.3 & 15892 & 0.18
& +0.77 & -7.35\% & 15.10 \\
ConCISE
& 84.1 & 853 & 9.86
& 83.3 & 2547 & 3.27
& 30.1 & 9315 & 0.32
& +0.83 & -51.5\% & 9.55 \\
MuTIS
& 85.4 & 895 & 9.54
& 84.6 & 1794 & 4.72
& 31.2 & 8976 & 0.35
& +2.07 & -56.7\% & 8.83 \\
TokenSqueeze
& -- & -- & --
& 83.2 & 2447 & 3.40
& 33.3 & 10672 & 0.31
& -- & -- & -- \\
\textbf{STACK}
& \textbf{87.4} & \textbf{684} & \textbf{12.78}
& \textbf{85.3} & \textbf{1783} & \textbf{4.78}
& \textbf{36.7} & \textbf{8752} & \textbf{0.42}
& \textbf{+4.80} & \textbf{-59.9\%} & \textbf{7.23} \\

\midrule
\multicolumn{13}{c}{\textit{DeepSeek-R1-Distill-Qwen-7B}} \\
% \midrule
Original
& 93.5 & 1478 & 6.32
& 91.2 & 4010 & 2.27
& 53.8 & 13178 & 0.41
& -- & -- & 13.29 \\
Prompt
& 95.1 & 1134 & 8.39
& 91.7 & 3654 & 2.51
& 53.4 & 11462 & 0.47
& +0.60 & -15.1\% & 12.47 \\
ConCISE
& 92.3 & 773 & 11.94
& 91.0 & 2016 & 4.51
& 48.5 & 7519 & 0.64
& -2.23 & -46.8\% & 8.60 \\
MuTIS
& 94.6 & 614 & 15.41
& 92.3 & 1887 & 4.89
& 54.7 & \textbf{6548} & \textbf{0.84}
& +1.03 & -53.9\% & 7.95 \\
TokenSqueeze
& -- & -- & --
& 92.8 & 1957 & 4.74
& 56.3 & 9079 & 0.62
& -- & -- & -- \\
\textbf{STACK}
& \textbf{95.7} & \textbf{563} & \textbf{17.00}
& \textbf{93.5} & \textbf{1733} & \textbf{5.39}
& \textbf{57.4} & 7274 & 0.79
& \textbf{+2.70} & \textbf{-54.5\%} & 6.73 \\

\bottomrule
\end{tabular}
\caption{Performance comparison of reasoning compression methods on three mathematical reasoning benchmarks.}
\label{tab:math_reasoning_compression}
\vspace{-12pt}
\end{table*}

\subsection{Loss function}
To explicitly model the compression benefit while maintaining the training stability of DPO, we introduce a reward-driven dynamic margin loss into the DPO, referred to as MDPO. Given an input $x$, along with the preferred sample $y_w$ and dispreferred sample $y_l$, the loss function is defined as: 
\vspace{-15pt}
% \begin{multline}
% \mathcal{L}_{\mathrm{MDPO}}(\pi_\theta; \pi_{\mathrm{ref}}) = - \mathbb{E}\left[\log \sigma\left(\beta \log \frac{\pi_\theta(y_w|x)}{\pi_{\mathrm{ref}}(y_w|x)} \\- \beta \log \frac{\pi_\theta(y_l|x)}{\pi_{\mathrm{ref}}(y_l|x)} - \Delta R \right ) \right ] \text{,}
% \end{multline}
\begin{multline}
\mathcal{L}_{\mathrm{MDPO}}(\pi_\theta; \pi_{\mathrm{ref}}) = - \mathbb{E} \Bigg[ 
\log \sigma \Bigg(
\beta \log \frac{\pi_\theta(y_w|x)}{\pi_{\mathrm{ref}}(y_w|x)}
\\ - \beta \log \frac{\pi_\theta(y_l|x)}{\pi_{\mathrm{ref}}(y_l|x)}
- \Delta R
\Bigg)
\Bigg] ,
\end{multline}
where $\sigma(\cdot)$ is the sigmoid function, $\beta$ controls the strength of the KL constraint relative to the reference policy $\pi_{\mathrm{ref}}$, and $\Delta R = \mathcal{R}_w - \mathcal{R}_l$ denotes the reward gap between samples induced by the reward function. It is important to note that if the compressed reasoning chain is correct, it is designated as the preferred sample; otherwise, the uncompressed correct reasoning chain is designated as the preferred sample.

% By incorporating $\Delta R$ as an adaptive discrimination margin, the loss function not only enables the model to learn preference direction but also explicitly requires the policy advantage to match the magnitude of actual reward improvement. This encourages more active exploration of efficient compression paths while ensuring factual correctness. When the reward difference is small, the loss naturally reduces to standard DPO, effectively suppressing excessive updates caused by noisy samples. Simultaneously, the log-ratio terms of the reference policy provide a stable anchor for training, preventing language degeneration or policy collapse when pursuing minimal output. This loss function effectively balances the incentive for compression with the reasoning quality and training stability.

By introducing $\Delta R$ as an adaptive discrimination margin, the loss function aligns the policy advantage with the magnitude of actual reward improvement, enabling the model to learn both preference direction and strength. 
% This design promotes efficient compression while preserving factual correctness. 
When the reward difference is small, the loss naturally degenerates to standard DPO, mitigating noisy or excessive updates. Meanwhile, the reference-policy log-ratio serves as a stabilizing anchor, preventing policy collapse or language degradation under aggressive compression. Overall, the loss achieves a principled trade-off between compression efficiency, reasoning quality, and training stability.

\begin{figure*}[t!]
\centering

% ================= Left: Accuracy =================
\begin{minipage}[t]{0.48\textwidth}
\centering
\begin{tikzpicture}

% ---- Left y-axis ----
\begin{axis}[
    width=\linewidth,
    height=5cm,
    % axis y line*=left,
    % axis x line=bottom,
    axis lines=box,
    xlabel={Training Step},
    ylabel={Accuracy (\%)},
    xmin=0, xmax=450,
    ymin=65, ymax=70,
    xtick={0,50,100,150,200,250,300,350,400,450},
    ytick={65,66,67,68,69,70},
    ymajorgrids=true,
    grid style=dashed,
    yticklabel style={text=c5},
    legend to name=sharedlegend,
    legend columns=2,
    legend style={draw=none}
]

\addplot[
    color=c5,
    mark=square*,
    line width=1.2pt,
    solid
] coordinates {
    (0,65)(50,65.8)(100,66.7)(150,67.5)(200,68.3)
    (250,69)(300,69.4)(350,69.6)(400,69.7)(450,69.8)
};

\end{axis}

% ---- Right y-axis ----
\begin{axis}[
    width=\linewidth,
    height=5cm,
    overlay,
    axis y line*=right,
    axis x line=none,
    xmin=0, xmax=450,
    ymin=78, ymax=83,
    ytick={78,79,80,81,82,83},
    yticklabel style={text=c1}
]

\addplot[
    color=c1,
    mark=triangle*,
    line width=1.2pt,
    solid
] coordinates {
    (0,79.5)(50,80.1)(100,80.7)(150,81.2)(200,81.6)
    (250,81.9)(300,82)(350,82.1)(400,82.15)(450,82.2)
};

\end{axis}

\end{tikzpicture}

% \centerline{(a) Accuracy}
\end{minipage}
\hfill
% ================= Right: Response Length =================
\begin{minipage}[t]{0.48\textwidth}
\centering
\begin{tikzpicture}
\begin{axis}[
    width=\linewidth,
    height=5cm,
    axis lines=box,
    xlabel={Training Step},
    ylabel={Response Length (tokens)},
    xmin=0, xmax=450,
    xtick={0,50,100,150,200,250,300,350,400,450},
    ymajorgrids=true,
    grid style=dashed
]

\addplot[color=c5, mark=square*, line width=1.2pt]
coordinates {
    (0,7979)(50,6650)(100,5600)(150,4800)
    (200,4300)(250,4000)(300,3850)
    (350,3750)(400,3700)(450,3739)
};
\addlegendentry{Qwen-1.5B}

\addplot[color=c1, mark=triangle*, line width=1.2pt]
coordinates {
    (0,6222)(50,5400)(100,4700)(150,4200)
    (200,3800)(250,3500)(300,3350)
    (350,3250)(400,3200)(450,3190)
};
\addlegendentry{Qwen-7B}

\end{axis}
\end{tikzpicture}

% \centerline{(b) Response Length}
\end{minipage}
\vspace{-3mm}

\caption{Average accuracy and response length of DeepSeek-R1-Distill-Qwen models across training steps.}
\label{fig:qwen_steps}

\vspace{-10pt}
\end{figure*}
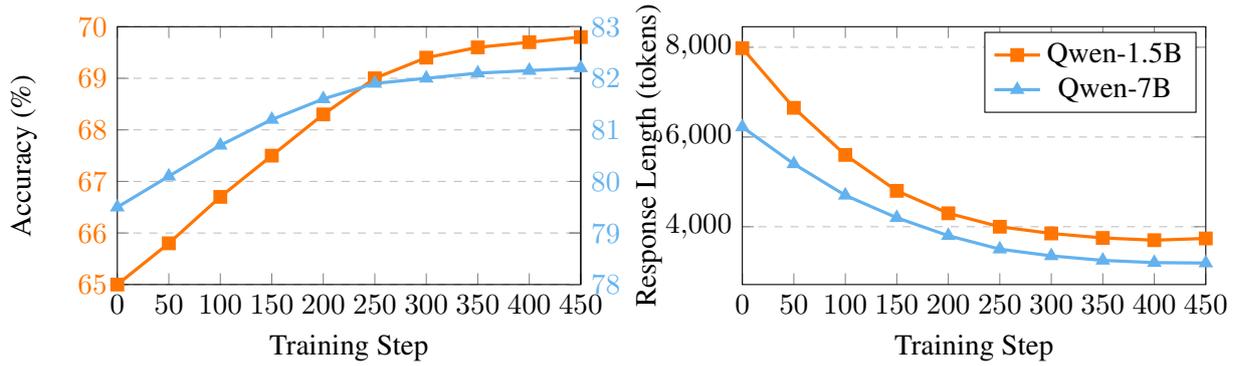

\section{Experiments}

\subsection{Setup}

We employ DeepSeek-R1-Distill-Qwen-1.5B and DeepSeek-R1-Distill-Qwen-7B~\cite{2025DeepSeek} as the base models and adopt STACK as the training framework. During training, the learning rate and batch size are set to 2e-6 and 64, respectively. The experiments are conducted using the PyTorch framework on 4×NVIDIA A100-80GB GPUs. Furthermore, to provide the model with external knowledge, following the approach of Search-o1~\cite{2025Search}, we utilize the Bing Web Search API and collect the top-5 retrieved results. 
Considering that LRMs lack the generation capability for short CoT and knowledge-guided logical reasoning, we perform a cold-start fine-tuning using annotated data to prevent mode collapse and training instability in subsequent DPO optimization.
Additional training details are provided in the appendix~\ref{sec:trainAppendix}.

 \textbf{Datasets}
 Given data requirements of the task and the base model's strong mathematical performance, we employ three math datasets of increasing difficulty: GSM8K~\cite{2021Training}, MATH500~\cite{2023lightman}, and AIME24. Accuracy (ACC), response length (Len) and inference latency (Lat) are used as evaluation metrics. We also adopt Token Efficiency ($\mathrm{TE} = \frac{Acc}{Len}*100$)  to assess the method's balance between accuracy and efficiency~\cite{yan2025long}.
 
 \textbf{Baselines}
 We compare STACK with the following representative Chain-of-Thought compression methods: \textbf{Prompt}: Injecting external knowledge into the input and using lightweight prompts to constrain the model's response length; \textbf{ConCISE}~\cite{qiao2025concise}: Constructing high-quality CoT compression data through confidence injection and early stopping, followed by fine-tuning LRMs using simple preference optimization; \textbf{MuTIS}~\cite{zhao2025mutis}: Performing multi-round dynamic interventions on reasoning trajectories during the online sampling phase and fine-tuning LLMs with original and compressed reasoning chains; \textbf{TokenSqueeze}~\cite{zhang2025tokensqueeze}: Constructing high-quality long-short sample pairs via adaptive depth selection and distribution-aligned language refinement, followed by model training using DPO-L.

\subsection{Performance Comparison on Reasoning Benchmarks}

Table~\ref{tab:math_reasoning_compression} reports the performance of various CoT compression methods across LRMs of different scales. Overall, STACK consistently achieves the best balance between reasoning efficiency and accuracy. Compared with the baseline, STACK reduces the average reasoning length by over 59.9\% while improving average accuracy by 4.8 points, yielding the highest token efficiency and the lowest inference latency. Moreover, the distribution of reasoning lengths across datasets with varying difficulty indicates that STACK enhances adaptive reasoning behavior. On the simpler GSM8K benchmark, the 1.5B model adaptively generates shorter reasoning chains (684 tokens on average).

% In contrast, prompt-based compression methods exhibit limited effectiveness, reducing inference latency by only about 1 second on average. This suggests that heuristic compression based solely on length constraints is insufficient for complex reasoning. Furthermore, despite incorporating external knowledge, such methods suffer a 1.1-point accuracy drop on AIME2024 with the 1.5B model. As reasoning chains grow longer, model generation becomes increasingly dominated by internal inference, thereby weakening the influence of external evidence. STACK overcomes these limitations through step-level fine-grained compression and explicit reinforcement of knowledge guidance at the output distribution level. Notably, this advantage persists on larger models: on the 7B model, STACK improves average accuracy by 2.7 points while reducing inference latency to 6.73s, compared with 7.23s on the 1.5B model. Nevertheless, the performance gap among different compression strategies remains substantial.

In contrast, prompt-based compression methods exhibit limited effectiveness, reducing inference latency by only approximately 1 second and failing to adequately support complex reasoning. Despite incorporating external knowledge, they incur a 1.1-point accuracy drop on AIME2024 using the 1.5B model.
% Even with external knowledge, they suffer a 1.1-point accuracy drop on AIME2024 with the 1.5B model, as long reasoning chains increasingly rely on internal inference, diminishing external guidance. 
As reasoning chains grow longer, model generation becomes increasingly dominated by internal inference, thereby weakening the influence of external evidence.
STACK addresses these issues via step-level fine-grained compression and explicit knowledge reinforcement at the output distribution. This advantage persists on larger models: on the 7B model, STACK improves average accuracy by 2.7 points while reducing inference latency to 6.73s (vs. 7.23s for the 1.5B model). Nevertheless, significant performance gaps remain among compression strategies.

ConCISE and TokenSqueeze significantly improve reasoning efficiency by constructing high-quality long–short sample pairs offline and fine-tuning models via preference optimization or contrastive learning. However, their statically constructed training data introduces a growing distribution shift as model capabilities evolve, which limits further performance gains. By contrast, STACK constructs contrastive samples online, enabling its compression strategy to adapt dynamically to variations across inputs and reasoning stages, resulting in superior overall performance.

MuTIS also adopts an online intervention paradigm and demonstrates strong performance across multiple settings. However, it merely focuses on structural-level compression strategy. In contrast, STACK continuously monitors step-level reasoning states and dynamically switches between self-prompted and knowledge-guided compression, enabling it to address both stage-specific  redundancy and reasoning bias. 
Experiments on challenging benchmarks such as MATH500 and AIME2024 show that this state-aware and knowledge-guided design substantially improves reasoning accuracy and token efficiency, underscoring the robustness of STACK in complex tasks. Concrete examples are provided in the appendix~\ref{sec:caseAppendix}.

\subsection{Training Dynamics across Model Scales}

% From the overall training dynamics, the combination of online contrastive CoT and reward-differential–driven optimization enables an adaptive balance between reasoning accuracy and representational compactness in accordance with model capacity. In the early training stage, the model primarily learns to remove repetitive verification and redundant reflection, leading to a rapid reduction in response length. As training progresses, compressed reasoning becomes increasingly constrained by accuracy requirements, and the model begins to improve factual correctness while preserving essential information, eventually reaching a convergence phase in which both length and accuracy stabilize.
% % This trajectory exhibits a characteristic pattern of “compression first, refinement later.”

% Models of different scales exhibit pronounced differences under this framework. The 1.5B model, with higher initial reasoning redundancy, achieves larger compression gains at early stages; however, constrained by limited capacity, its accuracy improvements are concentrated in the mid-training phase, and it remains relatively sensitive to reward variations in later stages. In contrast, the 7B model starts with more compact reasoning and thus has less compression headroom, but its stronger representational capacity enables more synchronized improvements in compression and accuracy. As a result, its training curve is smoother overall and converges within fewer training steps.

As shown in Figure~\ref{fig:qwen_steps}, the integration of online contrastive CoT construction with reward-difference–driven optimization enables an adaptive balance between reasoning accuracy and information density that aligns with model capacity. During early training, the model primarily eliminates repetitive verification and redundant reflection, resulting in a rapid reduction in response length. As training proceeds, compression becomes increasingly constrained by accuracy requirements, leading the model to preserve essential information and progressively improve factual correctness. The process eventually reaches a convergence mode characterized by stable reasoning length and accuracy.

Models of different scales exhibit pronounced differences under this framework. The 1.5B model, characterized by higher initial redundancy, achieves larger early-stage compression gains but exhibits accuracy improvements mainly in mid training and remains more sensitive to reward fluctuations in later stages due to limited capacity. In contrast, the 7B model starts with more compact reasoning and thus offers less compression headroom, yet its stronger representational capacity enables more coordinated improvements in compression and accuracy, yielding smoother training dynamics and faster convergence.

\subsection{Ablation study}

\begin{table}[t!]
\centering
\begin{tabular}{lcccc}
\toprule
\multirow{2}{*}{\textbf{Method}} & \multicolumn{2}{c}{\textbf{MATH500}} & \multicolumn{2}{c}{\textbf{AIME24}} \\
\cmidrule(lr){2-3} \cmidrule(lr){4-5}
 & \textbf{Acc} & \textbf{Len} & \textbf{Acc} & \textbf{Len} \\
\midrule
original & 91.2 & 4010 & 53.8 & 13178 \\
\midrule
\multicolumn{5}{l}{\textit{Ablating early stopping}} \\
end signal & 86.7 & 3374 & 31.7 & 10693 \\
consistency      & 93.3 & 1980 & 57.5 & 8157 \\
\midrule
\multicolumn{5}{l}{\textit{Ablating training paradigm}} \\
SFT       & 92.4 & 3016 & 56.7 & 9761 \\
SFT+DPO   & 93.2 & 1895 & 57.1 & 7586 \\
\textbf{STACK} & \textbf{93.5} & \textbf{1733} & \textbf{57.4} & \textbf{7274} \\
\bottomrule
\end{tabular}
\caption{Ablation study on DeepSeek-R1-Distill-Qwen-7B model.}
\vspace{-16pt}
\label{tab:ablation}
\end{table}

As shown in Table~\ref{tab:ablation}, in the ablation study on the early stopping mechanism, we replaced our stability-based early stopping strategy, which relies on answer distribution, with two alternative approaches: increasing the probability of generating an end-of-reasoning signal~\cite{2025Answer} and tracking answer consistency across consecutive steps~\cite{2025Early}. The results indicate that relying solely on the end-of-reasoning signal leads to premature termination of reasoning, significantly reducing both accuracy and output length, particularly on AIME24. In contrast, our early stopping mechanism demonstrates comparable reliability and stability to answer consistency while achieving greater compression of generated outputs. This highlights that our early stopping based on answer-level convergence not only ensures sufficiently thorough reasoning but also enhances the conciseness and efficiency of the outputs.

In the ablation study on training paradigms, training solely with teacher-labeled data (SFT) provides modest gains, allowing the model to begin adapting to short-chain reasoning and improving stability for subsequent training. Introducing direct preference optimization (SFT+DPO) further enhances the model's ability to produce concise reasoning. Notably, our approach (SFT+MDPO), by incorporating reward differentials into the DPO optimization objective, achieves a more effective balance between accuracy and output length, enabling the model to generate more concise reasoning chains while maintaining stable performance.

\begin{figure}[t!]
    \centering
    % 第一张子图
    \subfloat{%
        \begin{tikzpicture}[font=\Large, scale=0.4]
        \begin{axis}[
            axis lines=box,
            xlabel={$\mu$},
            ylabel={Accuracy (\%)},
            symbolic x coords={0.10,0.20,0.25,0.30,0.40,0.50},
            xtick=data,
            ymin=81.0, ymax=82.3,
            ymajorgrids=true,
            grid style=dashed,
            yticklabel style={text=c2},
            legend style={
                at={(0.02,0.02)},
                anchor=south west,
                draw=none,
                fill=none
            }
        ]
        \addplot[
            color=c2,
            mark=square*,
            line width=1.2pt,
            solid
        ] coordinates {
            (0.10,82.05)
            (0.20,82.15)
            (0.25,82.20)
            (0.30,82.10)
            (0.40,81.85)
            (0.50,81.40)
        };
        \end{axis}
        \begin{axis}[
            overlay,
            axis y line*=right,
            axis x line=none,
            symbolic x coords={0.10,0.20,0.25,0.30,0.40,0.50},
            ymin=2500, ymax=3500,
            yticklabel style={text=c4}
        ]
        \addplot[
            color=c4,
            mark=triangle*,
            line width=1.2pt,
            solid
        ] coordinates {
            (0.10,3500)
            (0.20,3300)
            (0.25,3190)
            (0.30,3050)
            (0.40,2850)
            (0.50,2550)
        };
        \end{axis}
        \end{tikzpicture}
        % \label{Fig-2b}
    }
    % \hfill
    \hspace*{0.028\textwidth}  % 强制间距
    % 第二张子图
    \subfloat{%
        \begin{tikzpicture}[font=\Large, scale=0.4]
        % ---- Left y-axis: Accuracy ----
        \begin{axis}[
            axis lines=box,
            xlabel={$\alpha$},
            symbolic x coords={0.1,0.3,0.5,0.7,1.0},
            xtick=data,
            ymin=81.0, ymax=82.3,
            ymajorgrids=true,
            grid style=dashed,
            yticklabel style={text=c2},
            legend style={
                at={(0.02,0.02)},
                anchor=south west,
                draw=none,
                fill=none
            }
        ]
        \addplot[
            color=c2,
            mark=square*,
            line width=1.2pt,
            solid
        ] coordinates {
            (0.1,82.0)
            (0.3,82.15)
            (0.5,82.2)
            (0.7,81.9)
            (1.0,81.2)
        };
        \end{axis}
        % ---- Right y-axis: Length ----
        \begin{axis}[
            overlay,
            ylabel={Response length},
            y label style={
                at={(axis description cs:1.43,0.5)},  % 横向在右侧外 5%，纵向居中
                anchor=center,                         % 锚点居中
            },
            axis y line*=right,
            axis x line=none,
            symbolic x coords={0.1,0.3,0.5,0.7,1.0},
            ymin=2300, ymax=3900,
            yticklabel style={text=c4}
        ]
        \addplot[
            color=c4,
            mark=triangle*,
            line width=1.2pt,
            solid
        ] coordinates {
            (0.1,3800)
            (0.3,3400)
            (0.5,3190)
            (0.7,2900)
            (1.0,2400)
        };
        \end{axis}
        \end{tikzpicture}
        % \label{Fig-3c}
    }
    \vspace{-3pt}
    \caption{Sensitivity analysis of hyperparameters $\mu$ and $\alpha$ on average accuracy and response length.}
    \label{fig:hyperFigure}
    \vspace{-16pt}
\end{figure}
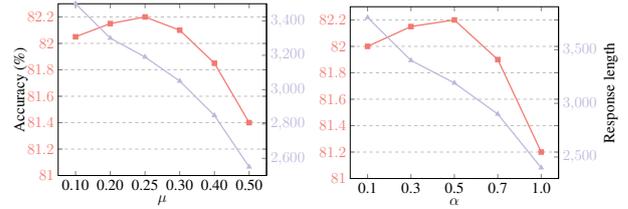

\subsection{Hyperparameter analysis}

Empirically, we find that the contrastive decoding coefficient $\mu$ and the compression reward weight $\alpha$ are crucial for balancing reasoning accuracy and chain compactness. Specifically, $\mu$ controls the magnitude of the penalty applied when the model's generation deviates from external evidence, while $\alpha$ determines the scaling factor of compression gains during training. 
Values that are too small make compression ineffective, while values that are too large cause premature truncation or reduced reasoning quality as shown in Figure~\ref{fig:hyperFigure}. Across different model scales, $\mu=0.25$ and $\alpha=0.5$ consistently achieve the best balance between preserving accuracy and compressing reasoning.

\section{Conclusion}
  
% In this work, we propose STACK, a State-Aware Reasoning Compression framework with Knowledge Guidance, which innovatively integrates a retrieval-augmented mechanism into the compression process and enables dynamic switching of online compression strategies to address diverse sources of redundancy in LRM reasoning. The introduction of early stopping mechanism based on answer-level convergence ensures thorough reasoning while further enhancing output conciseness. Moreover, by incorporating reward differences into the DPO optimization objective, STACK achieves a superior balance between accuracy and reasoning length, attaining state-of-the-art performance across different models and datasets.

In this work, we propose STACK, a State-Aware Reasoning Compression framework with Knowledge Guidance, which innovatively integrates a retrieval-augmented mechanism into the COT compression process and enables dynamic switching of online compression strategies to address diverse sources of redundancy at different steps along the reasoning chain. The introduction of early stopping mechanism based on answer-level convergence ensures thorough reasoning while further enhancing output conciseness. Moreover, by incorporating the reward differences into the DPO optimization objective, STACK achieves a superior balance between accuracy and reasoning length, attaining a state-of-the-art performance across different models and datasets.

\section{Limitations}

Despite the effectiveness of STACK in dynamically compressing reasoning chains and leveraging retrieval-augmented guidance, several limitations remain. First, the incorporation of online contrastive sampling and retrieval-based strategies introduces additional computational overhead, leading to increased training time and resource consumption. Second, the current retrieval mechanism does not guarantee fully accurate or relevant external knowledge, which may propagate suboptimal guidance during reasoning; joint optimization of the retrieval component and the model could further enhance performance. Finally, our framework currently relies solely on textual knowledge for external augmentation, while other modalities or external reasoning tools, such as structured databases, symbolic solvers, or multi-modal knowledge sources, represent promising directions for further extension and improvement.

\bibliography{custom}

\appendix

\section{Training Detail}
\label{sec:trainAppendix}

We train our models on the DeepScaleR dataset~\cite{luo2025deepscaler}, which comprises 40,000 mathematical problems sourced from AIME (1983–2023), AMC, Omni-Math~\cite{gao2024omni}, and STILL~\cite{min2024imitate}.
 In addition, considering that LRMs are not inherently equipped to generate short Chains of Thought (Short CoT) or perform knowledge-guided logical reasoning, directly applying DPO to an unadapted model would require the new policy $\pi_{\theta}$ to simultaneously learn the two challenging objectives of knowledge utilization and logical compression. This often results in a substantial divergence from the reference policy $\pi_{\text{ref}}$, leading to KL divergence explosion and training instability. To address this, we introduce a knowledge-guided cold-start stage designed to pre-align the reference policy distribution.

\textbf{Data Construction \& Mixing Strategy.} During the cold-start phase, we perform supervised fine-tuning (SFT) on the base model using a 5,000 sample dataset, composed of two parts mixed at an 80:20 ratio:
\begin{enumerate}
  \item Task-Specific Data (80\%).
   To enable the model to acquire compressed reasoning patterns conditioned on retrieved knowledge, we construct pairs of the form $(x, y_{\text{teacher}})$, where the input $x = [Q, I]$ consists of the question and retrieval context, and the output $y_{\text{teacher}}$ is a high-quality, knowledge-integrated, compressed Chain of Thought generated by a teacher model (GPT-4o).
   \item General Regularization Data (20\%).
   To prevent catastrophic forgetting caused by overfitting to specific compression patterns, we incorporate general QA and reasoning data (e.g., standard CoT), preserving the model's fundamental language and logical generalization capabilities.
\end{enumerate}

The primary objective of this phase is to construct a high-quality reference policy $\pi_{\text{ref}}$. By pre-adapting $\pi_{\text{ref}}$ to the conditional distribution $P(y \mid Q, I)$, we constrain the exploration space in the subsequent reinforcement learning phase from unguided trial-and-error to knowledge-grounded refinement, ensuring that the KL divergence constraint acts as a benign “language competence safeguard” rather than a barrier to effective knowledge injection.

\section{Analysis of Hesitation State}
\label{sec:hesitationAppendix}

\begin{figure*}
    \centering
    \includegraphics[width=1\textwidth]{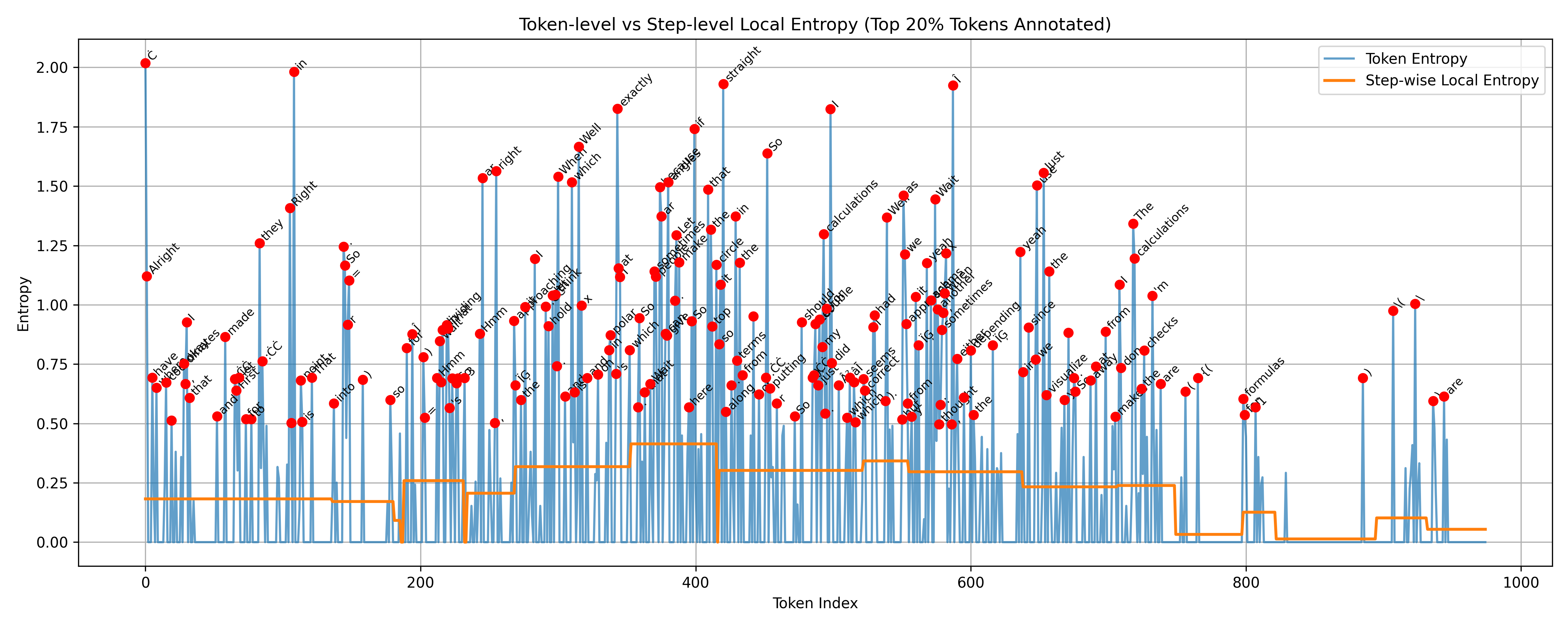}
    \caption{The distribution of token-level and step-level entropy.}
    \label{fig:hesitation_dis}
    \vspace{-5pt}
\end{figure*}

Due to the inherently iterative and self-reflective nature of long chain-of-thought reasoning, reflection tokens appear with high frequency across reasoning steps. Figure~\ref{fig:hesitation_dis} visualizes the corresponding distributions of token-level and step-level information entropy, illustrating the variation in model uncertainty throughout the reasoning process.

\begin{figure*}[htbp]
    \centering
    \small
    
    % 子图A: 平均检索次数
    \begin{subfigure}[b]{0.3\textwidth}
        \centering
        \begin{tikzpicture}
        \begin{axis}[
            ybar,
            bar width=20pt,
            width=0.95\textwidth,
            height=0.95\textwidth,
            ylabel={Avg Retrievals},
            symbolic x coords={Local Entropy, Single Reflection},
            xtick=data,
            xticklabel style={rotate=45, anchor=east, align=center, font=\footnotesize},
            enlarge x limits=0.4,
            ymin=0,
            ymax=10,
            ymajorgrids=true,
            nodes near coords,
            every node near coord/.style={
                font=\footnotesize,
                yshift=5pt
            },
            title={(a) Avg Retrievals},
            title style={yshift=5pt, font=\small\bfseries},
            ybar=0pt,
            bar shift=0pt,
        ]
        \addplot[fill=blue!50] coordinates {
            (Local Entropy,3)
            (Single Reflection,7)
        };
        \end{axis}
        \end{tikzpicture}
    \end{subfigure}
    \hfill
    % 子图B: 平均准确率
    \begin{subfigure}[b]{0.3\textwidth}
        \centering
        \begin{tikzpicture}
        \begin{axis}[
            ybar,
            bar width=20pt,
            width=0.95\textwidth,
            height=0.95\textwidth,
            ylabel={Avg Accuracy (\%)},
            symbolic x coords={Local Entropy, Single Reflection},
            xtick=data,
            xticklabel style={rotate=45, anchor=east, align=center, font=\footnotesize},
            enlarge x limits=0.4,
            ymin=80,
            ymax=84,
            ymajorgrids=true,
            nodes near coords,
            every node near coord/.style={
                font=\footnotesize,
                yshift=5pt
            },
            title={(b) Avg Accuracy},
            title style={yshift=5pt, font=\small\bfseries},
            ybar=0pt,
            bar shift=0pt,
        ]
        \addplot[fill=green!50] coordinates {
            (Local Entropy,82.2)
            (Single Reflection,81.5)
        };
        \end{axis}
        \end{tikzpicture}
    \end{subfigure}
    \hfill
    % 子图C: 平均响应长度
    \begin{subfigure}[b]{0.3\textwidth}
        \centering
        \begin{tikzpicture}
        \begin{axis}[
            ybar,
            bar width=20pt,
            width=0.95\textwidth,
            height=0.95\textwidth,
            ylabel={Avg Response Length},
            symbolic x coords={Local Entropy, Single Reflection},
            xtick=data,
            xticklabel style={rotate=45, anchor=east, align=center, font=\footnotesize},
            enlarge x limits=0.4,
            ymin=3000,
            ymax=4000,
            ymajorgrids=true,
            nodes near coords,
            every node near coord/.style={
                font=\footnotesize,
                yshift=5pt
            },
            title={(c) Avg Response Length},
            title style={yshift=5pt, font=\small\bfseries},
            ybar=0pt,
            bar shift=0pt,
        ]
        \addplot[fill=red!50] coordinates {
            (Local Entropy,3190)
            (Single Reflection,3650)
        };
        \end{axis}
        \end{tikzpicture}
    \end{subfigure}
    
    \caption{Comparison of local-entropy-based detection versus single reflection token-based detection across three metrics: (a) average number of retrievals, (b) average accuracy, and (c) average response length.}
    \label{fig:hesitation_bar}
\end{figure*}
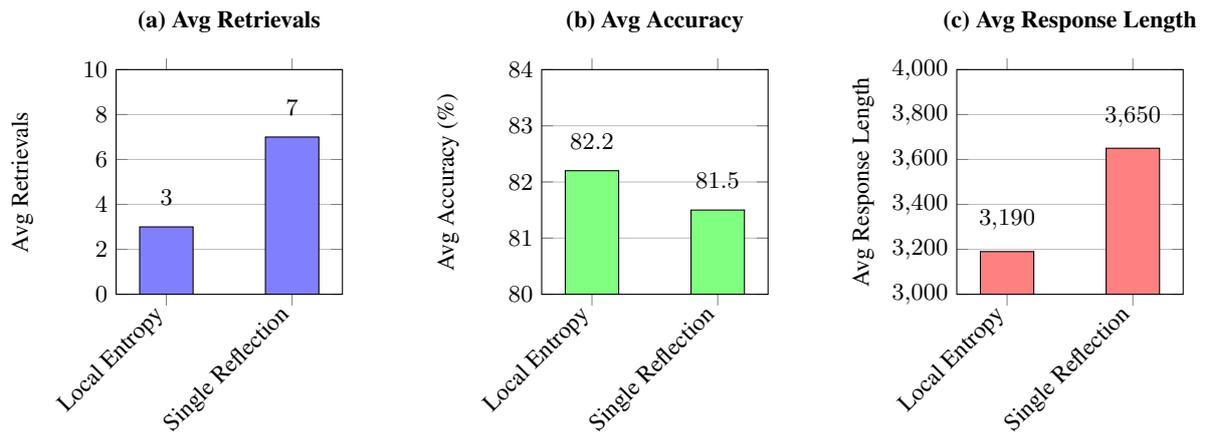

Compared to single reflection token-based detection, our local-entropy-based method significantly reduces unnecessary knowledge retrievals while maintaining or slightly improving reasoning accuracy. Specifically, as shown in figure~\ref{fig:hesitation_bar}, across long reasoning chains, entropy-based detection triggers only 3 retrievals per chain on average, achieving 82.2\% accuracy with a mean response length of 3,190 tokens. In contrast, relying on individual reflection tokens leads to 7 retrievals per chain, with slightly lower accuracy (81.5\%) and longer responses averaging 3,650 tokens.

In our opinion, this seemingly counterintuitive result arises because excessive external knowledge retrieval does not necessarily improve reasoning outcomes and can even degrade DPO training performance. Frequent retrievals introduce redundant or low-relevance information, which increases context length and disperses the model's attention across less critical tokens. During DPO optimization, the model is guided by reward differences between compressed and uncompressed chains; if the context is filled with repetitive or non-informative knowledge, the reward signal for effective compression becomes noisy. Consequently, the policy may struggle to correctly attribute which reasoning steps are truly valuable, leading to reduced accuracy and less efficient chain compression. By contrast, local-entropy-based detection selectively introduces external knowledge only when the model exhibits genuine hesitation, ensuring that retrieved evidence is relevant and timely. This targeted approach not only preserves reasoning fidelity but also stabilizes the training signal, resulting in more accurate and compact reasoning chains.

\section{Case Study}
\label{sec:caseAppendix}

The following tables present two representative cases comparing the original long CoT with the compressed CoT generated by STACK. From these examples, we observe that the excessive length of long CoT mainly stems from repeated verification under uncertain reasoning states (characterized by high local entropy) and redundant expressions that contribute little to task resolution.

By explicitly illustrating the compression process, the cases demonstrate how STACK performs step-wise state detection via local entropy, enabling the timely introduction of external knowledge to correctly steer the reasoning trajectory. This mechanism effectively mitigates reasoning bias, prevents error propagation, and suppresses ineffective reasoning. Furthermore, the answer-convergence–based early stopping mechanism truncates redundant verification once the solution stabilizes, further promoting reasoning-chain compression.

As a result, STACK achieves a favorable balance between conciseness and correctness, producing reasoning chains that are both compact and accurate.

\begin{figure*}[t]  % 让 tcolorbox 在顶部跨两栏
\begin{tcolorbox}[colback=gray!10,
                  colframe=black,
                  % width=\textwidth,
                  % arc=2mm, 
                  % auto outer arc,
                  title={Original output for Case 1}]
    % \small
\textbf{Question:} Convert the point $(0,3)$ in rectangular coordinates to polar coordinates.  Enter your answer in the form $(r,\theta)$, where $r > 0$ and $0 \le \theta < 2 \pi$. \\

\textbf{Output:} \\
Alright, so I need to convert the rectangular coordinate $(0,3)$ to polar coordinates. Hmm, polar coordinates are represented as $(r,\theta)$, where r is the distance from the origin to the point, and $\theta$ is the angle made with the positive x-axis. I remember that there are formulas to convert from rectangular $(x, y)$ to polar coordinates. \\

First, let me recall the formulas. I think $r$ is calculated using the Pythagorean theorem, so it should be the square root of ($x$ squared plus $y$ squared). So, $r = \sqrt{x^2 + y^2}$. Yeah, that sounds right because it's essentially finding the hypotenuse of a right triangle with sides $x$ and $y$. \\

Okay, so plugging in the values from the point $(0,3)$. Here, $x$ is $0$ and $y$ is $3$. Let me compute $r$. So $r = \sqrt{0^2 + 3^2}$ which is $\sqrt{0 + 9}$ which simplifies to $\sqrt{9}$. That's 3. So, $r$ is 3. That makes sense because the point is 3 units above the origin on the y-axis, so its distance from the origin is indeed 3. \\

Next, I need to find $\theta$, the angle. The formula for $\theta$ is arctangent of y over x, so $\theta = \arctan(\frac{y}{x})$. But wait, in this case, $x$ is 0. So, $\arctan(\frac{3}{0})$. Hmm, I remember that dividing by zero is undefined, I need to think about what angle $\theta$ gives a point on the positive y-axis. \\

Since $x$ is 0 and $y$ is 3, which is positive, the point lies on the positive y-axis. In polar coordinates, when a point is on the positive y-axis, what is the angle? I think it's $\frac{\pi}{2}$ radians because that's 90 degrees.
 Let me confirm. Starting from the positive x-axis, if we go counterclockwise 90 degrees, we reach the positive y-axis. So, $\theta$ should be $\frac{\pi}{2}$. \\

But let me make sure I'm not missing anything because sometimes when x is zero, the angle is either $\frac{\pi}{2}$  or $\frac{3\pi}{2}$  depending on the sign of y. Since y is positive here, it's definitely $\frac{\pi}{2}$. If y were negative, it would be $\frac{3\pi}{2}$ , but in this case, it's positive. \\

Let me visualize the point $(0, 3)$ on the Cartesian plane. It's straight up along the y-axis, so the angle from the positive x-axis is a right angle, which is $\frac{\pi}{2}$ radians. So, $\theta$ is $\frac{\pi}{2}$ . \\
Therefore, putting it all together, the polar coordinates are $(r,\theta) = (3,\frac{\pi}{2})$. \\

Wait, just to be thorough, sometimes when converting, especially with arctangent, you have to consider the quadrant in which the point lies. In this case, since the point is $(0,3)$, it's on the y-axis, not in any specific quadrant other than being on the axis itself. But since $\theta$ is between $\theta$ and $2\pi$, and our point is on the positive y-axis, $\theta$ should be $\frac{\pi}{2}$.\\

I think that's it. So, my final answer should be $(3, \frac{\pi}{2})$. Let me just double-check my calculations. \\
Calculating $r$: $\sqrt{0^2 + 3^2} = \sqrt{9} = 3$. That's correct. \\

Calculating $\theta$: $\arctan(\frac{3}{0})$. As $x$ is 0, and $y$ is positive, $\theta$ is $\frac{\pi}{2}$. That seems right. \\

No mistakes there. So, yeah, $(3, \frac{\pi}{2})$ is the polar coordinate for $(0, 3)$.\\

**Final Answer**The polar coordinates are $ \boxed { \left ( 3,\frac{\pi}{2}  \right ) }  $. \\

\textbf{Ground Truth:} $\left ( 3,\frac{\pi}{2}  \right ) $ 

\textbf{$\left [ \mathrm {True}    \right ] $}

\end{tcolorbox}
% \captionof{figure}{Original output for Case 1.}
\label{tcolorbox:originalOutput_case1}
\end{figure*}

\begin{figure*}[t]  % 让 tcolorbox 在顶部跨两栏
\begin{tcolorbox}[colback=gray!10,
                  colframe=black,
                  % width=\textwidth,
                  % arc=2mm, 
                  % auto outer arc,
                  title={STACK's compression process for Case 1}]
    % \small
\textbf{Question:} Convert the point $(0,3)$ in rectangular coordinates to polar coordinates.  Enter your answer in the form $(r,\theta)$, where $r > 0$ and $0 \le \theta < 2 \pi$. \\

\textbf{System Prompt:}\\
Please reason step by step, and put your final answer within \textbackslash boxed$\left \{  \right \} $ .

\textbf{Original Output:}
Okay, so I need to convert the rectangular coordinate $(0,3)$  to polar coordinates. Hmm, I remember that polar coordinates are represented as $(r,\theta)$, where $r$ is the radius or the distance from the origin, and $\theta$ is the angle made with the positive x-axis. 

\textbf{Local Entropy:} 0.2086 \\
\textbf{KL divergence:} 5.7693  \\

\textbf{Original Output:}
First, let me recall the formulas for converting rectangular coordinates $(x,y)$ to polar coordinates $(r,\theta)$. I think $r$ is calculated using the Pythagorean theorem, so $r = \sqrt{x^2 + y^2}$. And the angle $\theta$ can be found using the arctangent function, specifically $\theta = \arctan (\frac{y}{x}) $. But wait, I also remember that sometimes $\arctan (\frac{y}{x})$ isn't enough because it doesn't account for the correct quadrant. So, I might need to adjust the angle based on the signs of $x$ and $y$.

\textbf{Local Entropy:} 0.5049  \\
\textbf{KL divergence:} 0.2825  \\

\textbf{Knowledge-Guided Compression}\\
\textbf{System Prompt:}\\
Given the current reasoning and external knowledge:

Continue the reasoning ONLY if it adds new information.\\
Do NOT repeat, restate, or re-verify any known facts.\\
Use the external knowledge directly to reach the conclusion.\\

Once the answer is clear:
Put your final answer within \textbackslash boxed$\left \{  \right \} $  and STOP. \\

$\left [ \text{Current Reasoning} \right ] $ 
$\left \{ \text{current reasoning} \right \} $ \\

$\left [ \text{External Knowledge} \right ] $
Rectangular coordinates are represented as $(x, y)$, and polar coordinates are represented as $(r,\theta)$, where $r$ is the distance to the origin and $\theta$ is the angle with the x-axis. To convert from rectangular coordinates to polar coordinates, the formulas are as follows: $r = \sqrt{x^2 + y^2}$, $\theta = \arctan (\frac{y}{x}) $. \\

Assistant:\\

\textbf{Compression Output:}
To convert the rectangular coordinates $(0,3)$ to polar coordinates $(r,\theta)$, we can use the following formulas: \textbackslash n1. $r = \sqrt{x^2 + y^2}$  \textbackslash n2. $\theta = \arctan (\frac{y}{x}) $. Given the point $(0,3)$, we have $x = 0$ and $y = 3$.  \\

\textbf{Original Output:}
Alright, Let me compute r first. Plugging into the formula, $ r = \sqrt{0^2 + 3^2} = \sqrt{0 + 9} = \sqrt{9} = 3 $. Okay, that seems straightforward. So, $r$ is 3. 

\textbf{Local Entropy:} 0.3182 \\
\textbf{KL divergence:} 1.3271 \\

\textbf{Original Output:}
Okay, so I got $r$ as 3. Now I need to find the angle $\theta$. I remember that $\theta$ is the arctangent of $y$ over $x$. So $\theta = \arctan (\frac{y}{x}) $. But wait, I have to be careful here because $x$ is 0 in this case. So I need to think about what arctan of $\frac{3}{0}$ is. Hmm, 3 divided by 0 is undefined, so I need to figure out the angle that corresponds to 3 / 0 in the arctangent function.

\end{tcolorbox}
% \captionof{figure}{STACK's compression process for Case 1.}
\label{tcolorbox:stackCompression1_case1}
\end{figure*}

\begin{figure*}[t]  % 让 tcolorbox 在顶部跨两栏
\begin{tcolorbox}[colback=gray!10,
                  colframe=black,
                  % width=\textwidth,
                  % arc=2mm, 
                  % auto outer arc,
                  title={STACK's compression process for Case 1}]

\textbf{Local Entropy:} 0.7962\\
\textbf{KL divergence:} 0.6310 \\

\textbf{Knowledge-Guided Compression}\\
\textbf{System Prompt:}\\
Given the current reasoning and external knowledge:

Continue the reasoning ONLY if it adds new information.\\
Do NOT repeat, restate, or re-verify any known facts.\\
Use the external knowledge directly to reach the conclusion.\\

Once the answer is clear:
Put your final answer within \textbackslash boxed$\left \{  \right \} $  and STOP. \\

$\left [ \text{Current Reasoning} \right ] $
$\left \{ \text{current reasoning} \right \} $ \\

$\left [ \text{External Knowledge} \right ] $
The preconditions are as follows:  1. Take the origin of the rectangular coordinate system as the pole;  2. Use the non-negative half of the x-axis as the polar axis. \\

Assistant:\\

\textbf{Compression Output:}
Okay, so I got $r$ as 3. Now I need to find the angle $\theta$. I remember that $\theta$ is the arctangent of y over x. So $\theta = \arctan (\frac{y}{x}) $.
Since the angle from the positive x-axis to the point $(0, 3)$ is straight up along the y-axis, which is 90 degrees or $\frac{\pi}{2}$ radians. So $\theta$ is $\frac{\pi}{2}$.   \\

\textbf{Original Output:}
Okay, so putting it all together, the polar coordinates are $(3, \frac{\pi}{2})$. Let me just double-check my work to make sure I didn't make any mistakes. 

\textbf{Local Entropy:} 0.1484 \\
\textbf{KL divergence:} 0.3713 \\

\textbf{Original Output:}
Okay, so let me recap. The point is $(0, 3)$. To convert to polar coordinates, I use $r = \sqrt{x^2 + y^2}$. Plugging in $x=0$ and $y=3$, I get $r=3$. Then, for the angle, since $x$ is 0 and $y$ is positive, the angle is $\frac{\pi}{2}$. So the polar coordinates are $(3, \frac{\pi}{2})$. That seems correct. 

\textbf{Local Entropy:} 0.3764 \\
\textbf{KL divergence:} 0.0208 \\

\textbf{$\left [ \text{Early Stopping}  \right ] $}

\end{tcolorbox}
% \captionof{figure}{STACK's compression process for Case 1.}
\label{tcolorbox:stackCompression2_case1}
\end{figure*}

\begin{figure*}[t]  % 让 tcolorbox 在顶部跨两栏
\begin{tcolorbox}[colback=gray!10,
                  colframe=black,
                  % width=\textwidth,
                  % arc=2mm, 
                  % auto outer arc,
                  title={STACK's Output for Case 1}]
    % \small
\textbf{Question:} Convert the point $(0,3)$ in rectangular coordinates to polar coordinates.  Enter your answer in the form $(r,\theta)$, where $r > 0$ and $0 \le \theta < 2 \pi$. \\

\textbf{External Knowledge:} Rectangular coordinates are represented as $(x, y)$, and polar coordinates are represented as $(r,\theta)$, where $r$ is the distance to the origin and $\theta$ is the angle with the x-axis. To convert from rectangular coordinates to polar coordinates, the formulas are as follows: $r = \sqrt{x^2 + y^2}$, $\theta = \arctan (\frac{y}{x}) $. The preconditions are as follows:  1. Take the origin of the rectangular coordinate system as the pole;  2. Use the non-negative half of the x-axis as the polar axis.

\textbf{Output:} \\
To convert the rectangular coordinates $(0,3)$ to polar coordinates $(r,\theta)$, we can use the following formulas: \textbackslash n1. $r = \sqrt{x^2 + y^2}$  \textbackslash n2. $\theta = \arctan (\frac{y}{x}) $. Given the point $(0,3)$, we have $x = 0$ and $y = 3$. \\

First, let's calculate $r$: $ r = \sqrt{0^2 + 3^2} = \sqrt{0 + 9} = \sqrt{9} = 3 $. So, $r = 3$. \\

Next, let's calculate $\theta$. Since $x = 0$, we have a point on the y-axis. The arctangent function $\arctan\left(\frac{y}{x}\right)$ is undefined when $x = 0$ (since division by zero is not allowed). However, we can determine the angle based on the position of the point. \\

When $x = 0$ and $y > 0$, the point is on the positive y-axis. The angle $\theta$ for this position is $\frac{\pi}{2}$ radians (90 degrees). Therefore, $\theta = \frac{\pi}{2}$. \\

Putting it all together, the polar coordinates are $(3, \frac{\pi}{2})$. \\

**Final Answer**The polar coordinates of the point $(0,3)$ are $\boxed { \left ( 3,\frac{\pi}{2}  \right ) }  $.

\textbf{Ground Truth:} $\left ( 3,\frac{\pi}{2}  \right ) $ 

\textbf{$\left [ \mathrm {True}    \right ] $}

\end{tcolorbox}
% \captionof{figure}{STACK's Output for Case 1.}
\label{tcolorbox:stackOutput_case1}
\end{figure*}

\begin{figure*}[t]  % 让 tcolorbox 在顶部跨两栏
\begin{tcolorbox}[colback=gray!10,
                  colframe=black,
                  % width=\textwidth,
                  % arc=2mm, 
                  % auto outer arc,
                  title={Original Output for Case 2}]
    % \small
\textbf{Question:} The expression $2\cdot 3 \cdot 4\cdot 5+1$ is equal to 121, since multiplication is carried out before addition. However, we can obtain values other than 121 for this expression if we are allowed to change it by inserting parentheses. For example, we can obtain 144 by writing $2\cdot (3\cdot 4) \cdot (5+1) = 144$. In total, how many values can be obtained from the expression $2\cdot 3\cdot 4 \cdot 5 + 1$ by inserting parentheses? (Note that rearranging terms is not allowed, only inserting parentheses). \\

\textbf{Output:}\\
\dots \\
Wait, so that's another way to get 144.

So, 144 is achievable in multiple ways.

But earlier, I had 36 as another value when I incorrectly grouped 3*4 and 5+1 with 2. But that might not be a valid parenthesization because it changes the structure.

So, perhaps 36 is not a valid result.

Wait, let me check again.

If I group \(2 \cdot (3 \cdot 4) + (5 + 1)\), that would be 2*12 + 6 = 24 + 6 = 30. But in the original expression, it's \(2 \cdot 3 \cdot 4 \cdot 5 + 1\), so this grouping would require changing the order of operations, which isn't allowed.

So, that's not a valid parenthesization.

Therefore, only groupings that change the order of operations by grouping adjacent operations are allowed.

Therefore, the different groupings that result in different values are:

1. Grouping the addition with 5: 144.

2. Grouping the addition with 4 and 5: 126.

3. Grouping the addition with 3,4,5: 122.

4. Grouping the addition with 2,3,4,5: 121.

So, that's four different values.

Wait, but in the first grouping, when I grouped the addition with 5, I had 144, but when I grouped the addition with 4 and 5, I also got 144. So, that suggests that grouping the addition with 5 and grouping it with 4 and 5 both result in 144. So, it's the same value.

Similarly, grouping the addition with 3,4,5 gives 122, which is different.

Grouping the addition with 2,3,4,5 gives 121.

So, in total, we have 121, 122, 126, and 144.

Therefore, four different values. \\

\dots \\

Wait, but if you group \(3 + 4\), it's not part of a multiplication.

So, for example, \(2 \cdot (3 + 4) \cdot 5 + 1\). But in the original expression, it's \(2 \cdot 3 \cdot 4 \cdot 5 + 1\). So, grouping 3 + 4 would require changing the structure, which isn't allowed.

Therefore, that's not a valid parenthesization.

So, I think I can conclude that the different values are 121, 122, 126, and 144.

Therefore, the total number of different values is 4.

But in reality, in the original expression, it's left-associative, so 2 * 3 * 4 * 5 is equivalent to (((2 * 3) * 4) * 5). So, to group 3 * (4 + 5), you have to change the order of operations, which is allowed by inserting parentheses.

Therefore, is 55 a valid result?

\end{tcolorbox}
% \captionof{figure}{STACK's Output for Case 2.}
\label{tcolorbox:originalOutput1_case2}
\end{figure*}

\begin{figure*}[t]  % 让 tcolorbox 在顶部跨两栏
\begin{tcolorbox}[colback=gray!10,
                  colframe=black,
                  % width=\textwidth,
                  % arc=2mm, 
                  % auto outer arc,
                  title={Original Output for Case 2}]

Compute: 3 + 4 = 7, 5 + 1 = 6, 2 * 7 * 6 = 84. But in the original expression, it's a linear chain, so grouping 3 + 4 and 5 + 1 is not allowed. Therefore, 84 is invalid.

\dots\\

Wait, when I grouped the addition with 3,4,5, I got 122.But is 122 achievable in another way? Wait, if I group 3,4,5,1, it's 121. Wait, but 122 is unique. Similarly, 55 is unique. So, yes, five different values. Therefore, I think the answer is 5.

**Final Answer**
The number of different values is \boxed{5}.
</think>

To determine how many different values can be obtained by inserting parentheses into the expression \(2 \cdot 3 \cdot 4 \cdot 5 + 1\), we need to consider different ways of parenthesizing the multiplication and addition operations.

1. **Original Expression**: The expression is evaluated left to right without any parentheses.
   \[
   2 \cdot 3 \cdot 4 \cdot 5 + 1 = 120 + 1 = 121
   \]

2. **Grouping the Addition with 5**: Insert parentheses around 4 and 5.
   \[
   2 \cdot 3 \cdot (4 \cdot 5) + 1 = 2 \cdot 3 \cdot 20 + 1 = 120 + 1 = 121
   \]
   This gives the same result as the original expression.

3. **Grouping the Addition with 4 and 5**: Insert parentheses around 3, 4, and 5.
   \[
   2 \cdot (3 \cdot 4 \cdot 5) + 1 = 2 \cdot 60 + 1 = 120 + 1 = 121
   \]
   This also gives the same result as the original expression.

4. **Grouping the Addition with 3, 4, and 5**: Insert parentheses around 3, 4, and 5.
   \[
   2 \cdot (3 \cdot (4 \cdot 5)) + 1 = 2 \cdot 60 + 1 = 120 + 1 = 121
   \]
   This again gives the same result as the original expression.

5. **Grouping the Addition with 4 and 5**: Insert parentheses around 4 and 5.
   \[
   2 \cdot (3 \cdot (4 + 5)) + 1 = 2 \cdot 27 + 1 = 54 + 1 = 55
   \]

6. **Grouping the Addition with 3, 4, and 5**: Insert parentheses around 3, 4, and 5.
   \[
   (2 \cdot 3) \cdot (4 \cdot 5) + 1 = 6 \cdot 20 + 1 = 120 + 1 = 121
   \]

7. **Grouping theAddition with 5**: Insert parentheses around 5.
   \[
   (2 \cdot 3 \cdot 4) \cdot (5 + 1) = 24 \cdot 6 = 144
   \]

8. **Grouping theAddition with 3, 4, and 5**: Insert parentheses around 3, 4, and 5.
   \[
   ((2 \cdot 3) \cdot 4) \cdot (5 + 1) = 24 \cdot 6 = 144
   \]

After evaluating all possible groupings, the different values obtained are 55, 121, 122, 126, and 144. Therefore, the number of different values is:

\[
\boxed{5}
\]

\textbf{Ground Truth:} $4$ 

\textbf{$\left [ \mathrm {False}    \right ] $}

\end{tcolorbox}
% \captionof{figure}{STACK's Output for Case 2.}
\label{tcolorbox:originalOutput2_case2}
\end{figure*}

\begin{figure*}[t]  % 让 tcolorbox 在顶部跨两栏
\begin{tcolorbox}[colback=gray!10,
                  colframe=black,
                  % width=\textwidth,
                  % arc=2mm, 
                  % auto outer arc,
                  title={STACK's Output for Case 2}]
    % \small
\textbf{Question:} The expression $2\cdot 3 \cdot 4\cdot 5+1$ is equal to 121, since multiplication is carried out before addition. However, we can obtain values other than 121 for this expression if we are allowed to change it by inserting parentheses. For example, we can obtain 144 by writing $2\cdot (3\cdot 4) \cdot (5+1) = 144$. In total, how many values can be obtained from the expression $2\cdot 3\cdot 4 \cdot 5 + 1$ by inserting parentheses? (Note that rearranging terms is not allowed, only inserting parentheses). \\

\textbf{External Knowledge:}  Step 1: Initial Calculation
Calculate the initial expression without any parentheses: \(2 \cdot 3 \cdot 4 \cdot 5 + 1\).
Step 2: Possible Parentheses Placements
Identify the possible locations where parentheses can be inserted to change the order of operations.
Step 3: Evaluate Parentheses Configurations
Calculate the expression for each valid placement of parentheses.
Step 4: Count Unique Results
Determine the total number of unique values obtained by inserting parentheses. \\

\textbf{Output:}\\
The expression \(2 \cdot 3 \cdot 4 \cdot 5 + 1\) is equal to 121 when evaluated from left to right due to the order of operations. However, by inserting parentheses, we can obtain different values. 

We need to consider all possible ways to insert parentheses in the expression \(2 \cdot 3 \cdot 4 \cdot 5 + 1\) without rearranging the terms. The key is to explore different groupings of the operations.

1. **Default evaluation without parentheses:**
   \[
   2 \cdot 3 \cdot 4 \cdot 5 + 1 = 120 + 1 = 121
   \]

2. **Grouping the first two terms and the rest:**
   \[
   (2 \cdot 3) \cdot 4 \cdot 5 + 1 = 6 \cdot 4 \cdot 5 + 1 = 120 + 1 = 121
   \]
   This does not change the result.

3. **Grouping the first three terms and the last term:**
   \[
   (2 \cdot 3 \cdot 4) \cdot 5 + 1 = 24 \cdot 5 + 1 = 120 + 1 = 121
   \]

4. **Grouping the last two terms first:**
   \[
   2 \cdot 3 \cdot (4 \cdot 5) + 1 = 2 \cdot 3 \cdot 20 + 1 = 120 + 1 = 121
   \]
   This does not change the result.

5. **Grouping the middle terms first:**
   \[
   2 \cdot (3 \cdot 4) \cdot (5 + 1) = 2 \cdot 12 \cdot 6 = 144
   \]

6. **Grouping the last term with the addition:**
   \[
   2 \cdot (3 \cdot 4 \cdot 5 + 1) = 2 \cdot (60 + 1) = 2 \cdot 61 = 122
   \]

7. **Grouping the middle terms and then adding:**
   \[
   2 \cdot (3 \cdot (4 \cdot 5 + 1)) = 2 \cdot (3 \cdot 21) = 2 \cdot 63 = 126
   \]

By evaluating all possible groupings, we find the distinct values: 121, 144, 126, and 122. 
Thus, the total number of distinct values is \(\boxed{4}\). \\

\textbf{Ground Truth:} $4$ 

\textbf{$\left [ \mathrm {True}    \right ] $}

\end{tcolorbox}
% \captionof{figure}{STACK's Output for Case 2.}
\label{tcolorbox:stackOutput_case2}
\end{figure*}

\end{document}